\documentclass{vgtc}     

\usepackage{amsmath}
\usepackage{float}
\usepackage{pbalance}

\onlineid{8038}

\vgtccategory{Research}

\vgtcpapertype{system}

\title{FlockGPT: Guiding UAV Flocking with Linguistic Orchestration}

\author{%
  \authororcid{Artem Lykov}{0000-0001-6119-2366},
 Sausar Karaf, Mikhail Martynov, Valerii Serpiva, \\
 \authororcid{Aleksey Fedoseev}{0000-0003-2506-9111}, Mikhail Konenkov and \authororcid{Dzmitry Tsetserukou}{0000-0001-8055-5345}
}

\authorfooter{
  \item
  	Artem Lykov is with Skoltech.
  	E-mail: Artem.Lykov@skoltech.ru

}

\abstract{%
This article presents the world's first rapid drone flocking control using natural language through generative AI. The described approach enables the intuitive orchestration of a flock of any size to achieve the desired geometry. The key feature of the method is the development of a new interface based on Large Language Models to communicate with the user and to generate the target geometry descriptions. Users can interactively modify or provide comments during the construction of the flock geometry model. By combining flocking technology and defining the target surface using a signed distance function, smooth and adaptive movement of the drone swarm between target states is achieved. 

Our user study on FlockGPT confirmed a high level of intuitive control over drone flocking by users. Subjects who had never previously controlled a swarm of drones were able to construct complex figures in just a few iterations and were able to accurately distinguish the formed swarm drone figures. The results revealed a high recognition rate for six different geometric patterns generated through the LLM-based interface and performed by a simulated drone flock (mean of 80\% with a maximum of 93\% for cube and tetrahedron patterns). Users commented on low temporal demand (19.2 score in NASA-TLX), high performance (26 score in NASA-TLX), attractiveness (1.94 UEQ score), and hedonic quality (1.81 UEQ score) of the developed system. The FlockGPT demo code repository can be found at: \textbf{link omitted for submission}
}


\teaser{
  \centering
  \includegraphics[width=1\linewidth, alt={}]{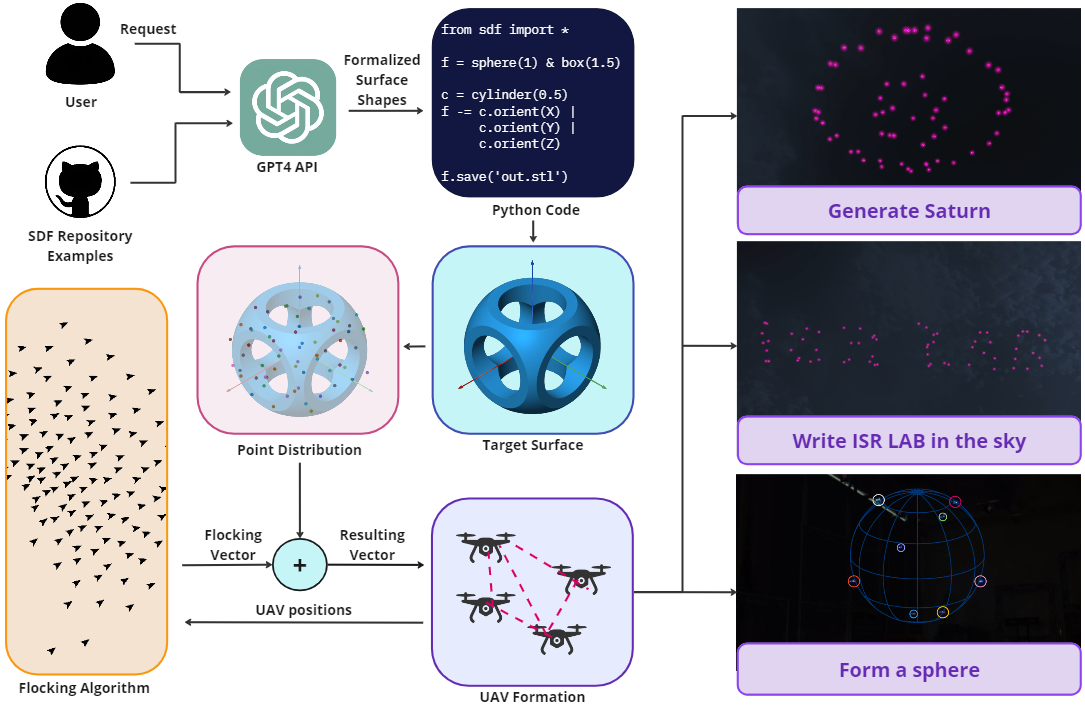}
  \caption{FlockGPT architecture. Left: Structure of LLM-based drone flocking control. Right: Visualization of drone flocking in VR and in the real environment.}
  \label{fig:system_architecture}
}

\graphicspath{{figs/}{figures/}{pictures/}{images/}{./}} 

\usepackage{tabu}                      
\usepackage{booktabs}                  
\usepackage{lipsum}                    
\usepackage{mwe}                       

\usepackage{mathptmx}                  

\begin{document}

\firstsection{Introduction}
\maketitle

The field of cognitive robotics is currently experiencing explosive growth, with a multitude of different embodiments emerging on various platforms. In mobile robots and manipulators, the gold standard is set by Google DeepMind with the development of PaLM-E \cite{driess2023palm}, RT1 \cite{brohan2022rt}, and RT2 \cite{brohan2023rt}. These robots demonstrate outstanding success in reasoning and interacting with real-world objects. In the field of cognitive four-legged robots, notable works include CognitiveDog by Lykov et al. \cite{lykov2024cognitivedog} leveraging natural design for enhanced real-world mobility. Additionally, research is underway on creating the humanoid robots, which offer the advantage of seamless integration into human environments, pursued by companies, e.g. Tesla \cite{Tesla_bot}, Agility Robotics \cite{DIGIT}, and OpenAI \cite{figure2024}.

However, such approaches tend to be robot-centric, and their scalability is associated with a proportional increase in required resources. Moreover, optimization of control approaches of multi-agent systems typically involves one model sequentially generating behaviors or paths for each agent in the system \cite{kannan2023smart, lykov2023llmmars, jiao2023swarmgpt}. Several systems combine human-robot interface and machine learning algorithms for adaptive control over drone swarms \cite{dorzhieva_2022}. However, this approach is also not scalable for real-time behavior generation with large-scale fleets. Thus, as the number of agents increases, a trade-off exists between response time and computational resources.

One of the most popular and widely applied types of multi-agent robotic systems is drone swarms. Whether used for terrain inspection or drone shows, these swarms can be quite large, comprising thousands of unmanned aerial vehicles (UAVs) in a single swarm. Timely generation of plans or paths for each of them would pose a significant technological challenge. Meanwhile, the control optimization of these drone formations can be significantly enhanced through the application of generative AI.

In this paper, we propose a novel approach for controlling drone flocks using generative AI, regardless of swarm size. We achieve this by generating target surfaces that the swarm should form and employing flocking to distribute drones across these surfaces. This approach utilizes a large language model to process natural language input once for the entire swarm, determining the necessary direction and speed of movement for each drone at any given point in space. Additional features include editing the swarm patterns interactively through refinements and communicating with an AI that responds textually, guiding the swarm of drones. The proposed approach is implemented in Unity and Gazebo simulations for testing on flocks with a large number of drones, as well as on real drone swarms. A conducted user study demonstrated a high level of adaptability of FlockGPT and the intelligence of control for novice users.

\section{Related Works}

\begin{table}[tb]
  \caption{%
        Results of the LLM SDF generation test.%
  }
  \label{tab:sdf_python}
  \scriptsize%
  \centering%
  \begin{tabu}{%
          X[0.5,l]%
                *{3}{X[0.25,c]}%
        }
        \toprule
        Model & Ability to generate SDF & Ability to generate the correct geometry & Ability to iteratively edit geometry \\
        \midrule
        llama-2-70b-chat & 25\% & 20\% & 10\% \\
        mixtral-8x7b-instruct-v0.1 & 40\% & 50\% & 30\% \\
        starling-lm-7b-beta & 55\% & 50\% & 30\% \\
        mistral-large-2402 & 60\% & 60\% & 50\% \\
        claude-3-sonnet-20240229 & 60\% & 60\% & 60\% \\
        qwen1.5-72b-chat & 70\% & 60\% & 60\% \\
        gpt-3.5-turbo-0613 & 75\% & 80\% & 60\% \\
        gemini-pro-dev-api & 70\% & 70\% & 70\% \\
        \textbf{gpt-4-1106-preview} & \textbf{90\%} & \textbf{90\%} & \textbf{90\%} \\
        \bottomrule
  \end{tabu}%
\end{table}

This study builds upon recent developments in the field of generative AI and enhanced swarm drone control. Let us consider some key contributions.

\textbf{Signed Distance Function:} To define the movement of all UAVs in formation, we have decided to leverage the Signed Distance Function (SDF) described in \cite{osher2006level}, which determines the distance from a given point in space to the nearest surface and the direction to it. This enables us to determine the direction of movement for any point in space to reach the surface. The functionality of generating SDF surfaces, akin to Computer-Aided Design (CAD), is implemented as an open-source Python library \cite{fogleman_sdf}. This repository provides everything necessary to create complex objects from primitives, to perform geometric transformations, and to generate 3D text. Thus, it allows us to specify target surfaces in a convenient Python code format and to obtain the flight direction of UAVs to place them on these surfaces based on their known positions in space.

\begin{figure}[h!]
  \centering
  \includegraphics[width=\columnwidth]{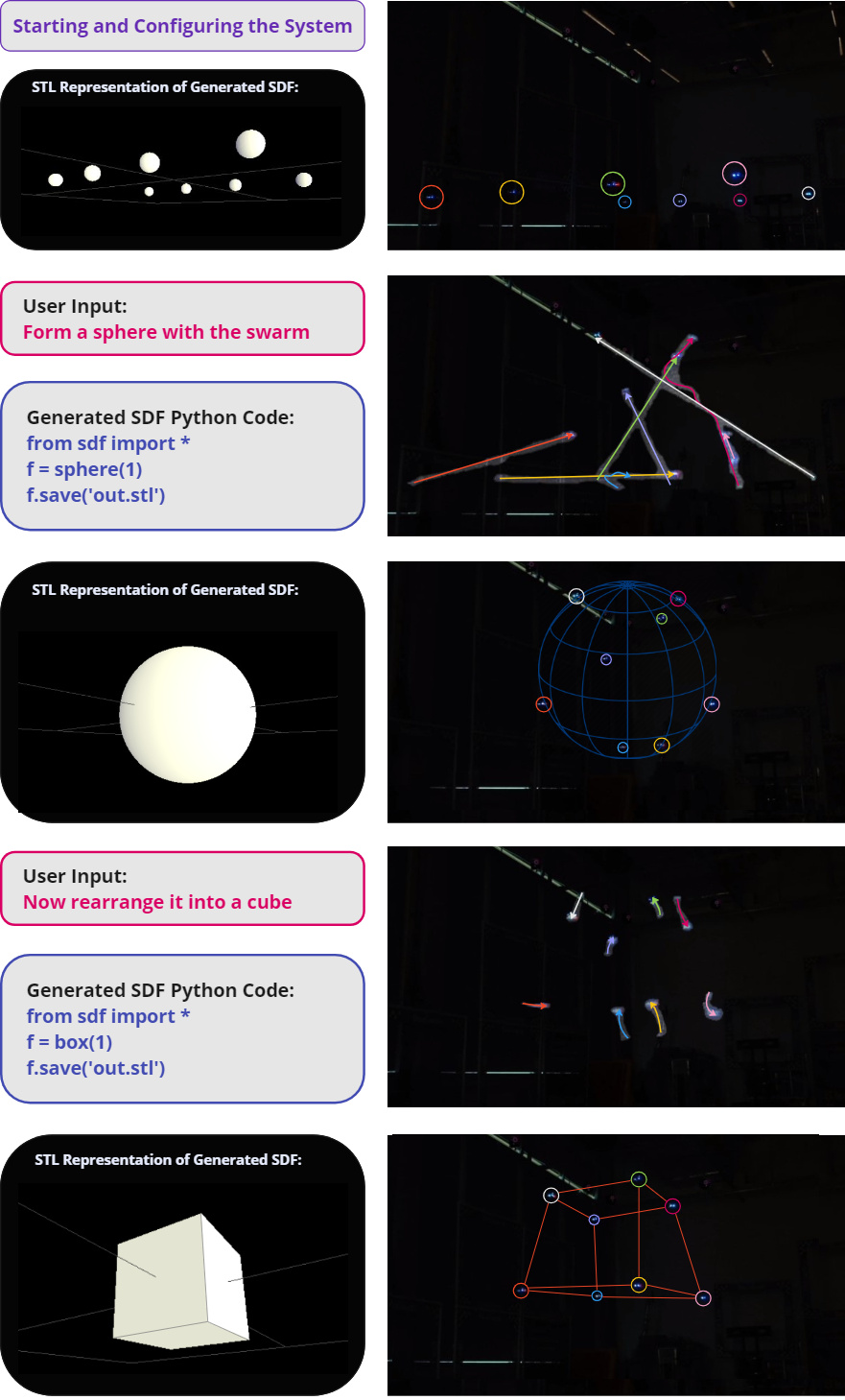}
  \caption{Demonstration of FlockGPT setup with a swarm of 8 Crazyflie 2.1 drones.}
  \label{fig:real_setup}
\end{figure}

\textbf{SDF Generation with LLM:} The choice of representing SDF in Python code format was deliberate. Most modern commercial Large Language Models, i.e. GPT3.5 \cite{lib:openai2022introducing}, GPT4 \cite{openai2023gpt4}, Claude3 \cite{Anthropic2024}, and Gemini Pro \cite{Gemini2023}, possess the ability to generate Python code. However, constructing complex composite figures in 3D space from descriptions requires additional emergent capabilities. To select a model, a test was conducted on the Chatbot Arena \cite{chiang2024chatbot}, where various commercial and open-source models were tasked with creating different figures based on user requests, using examples from the aforementioned SDF library as context. Without fine-tuning, open models, such as LLaMa2 \cite{lib:touvron2023llama}, Mistral \cite{jiang2023mistral}, Qwen \cite{bai2023qwentech}, and Starling \cite{starling2023} showed varying performance but were unstable in results. Meanwhile, the aforementioned commercial models demonstrated high performance. The comprehensive results of testing the ability of LLMs to generate SDFs in Python, to generate the correct geometry, and to iteratively edit geometry are presented in Table \ref{tab:sdf_python}. Due to the promising test results and ease of use, GPT4 from OpenAI was chosen as the base model for the project. The interaction was facilitated using the OpenAI API.

\textbf{Floaking in Swarm of Drones:} Once we have the direction for each drone towards the target surface, we aim to ensure that they all converge onto it by applying control methods. However, it is not enough for UAVs to be on the surface; they must be uniformly distributed across it and move as a cohesive flow, avoiding collisions. To achieve this, we employ a flocking algorithm based on the optimized autonomous drones in a confined environment \cite{flock2018optimized}. The presented tunable distributed flocking model for a large group of autonomous flying robots maintained stable, collision-free collective motion in a closed space with or without obstacles, exhibiting rich dynamics of motion, with a variety of emergent collective motion patterns. Moreover, elements of potential fields enable to control drones, to avoid obstacles, and to maintain the necessary formation \cite{sun2020path}. The Artificial Potential Field (AFP) method reacts swiftly to obstacles and it is well-suited for use with UAVs in dynamic environments \cite{hao2023uav}. It is also suitable for a wide range of drone models and flock sizes.

\section{System Architecture of FlockGPT}

The architecture of FlockGPT is illustrated in Fig. \ref{fig:system_architecture}. Interaction between the user and the swarm occurs entirely in natural language. In any explicit or implicit manner, the user specifies the shape of the swarm they expect to receive. The dialogue with the GPT4 model begins through the OpenAI API. Initially, we need to explain to the model the principles of the SDF library operation. To achieve this, based on the content of the README repository on GitHub, a set of examples for creating the predefined primitives and geometry editing tools available in the library were prepared. It has been proven that few-shot learning is an excellent and reliable tool for transferring model capabilities that it did not possess before \cite{fewshot}. To take advantage of step-by-step reasoning, the examples were supplemented with comprehensive comments that do not affect the code but help the model to more clearly understand the nuances of figure construction both in the examples and during response generation.

Once the system prompt is passed to the model, it also receives a natural language request from the user to generate a target surface. The model's output may contain additional comments to assist the model itself and include Python code for constructing the SDF of the Target Surface. Superfluous text is discarded using regular expression parsing, and we obtain functional code. This code contains the SDF as a Python function, which, for any point in space, determines the direction and distance to the target surface. This vector is then combined with a flocking algorithm and its associated rules. The implemented model simulates the behavior of a flock, in which agents adjust their speeds depending on the position and speeds of the nearby agents. Therefore, the combination of flocking behavior and attraction vectors to the generated surface enables the drones to be distributed over the figure for visualization purposes. The proposed expansion adds the ability for drones to be evenly distributed over the target surface.

\begin{figure}[tb]
  \centering
  \includegraphics[width=0.9\columnwidth]{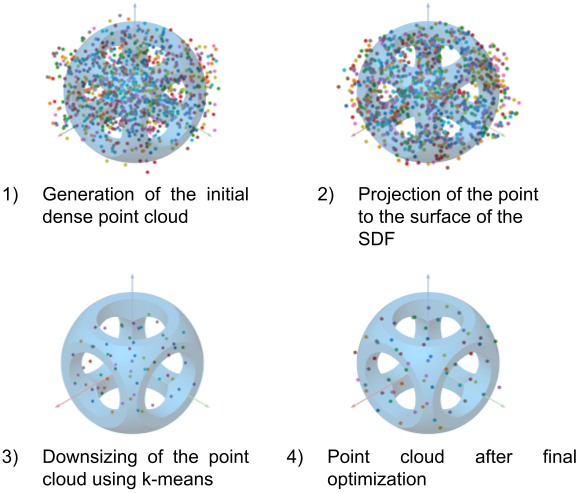}
  \caption{Point distribution visualization.}
  \label{fig:point_distribution}
\end{figure}

Subsequently, resulting vectors are transferred via the ROS interface both to the Crazyflie framework on the main computer and to the Unity simulation environment. The 3D velocity vector directs the drone towards its designated positions. The drone's position is continuously recorded by a motion capture system. To ensure precise tracking of the drones, we used the Vicon motion capture system, equipped with 14 Vantage V5 IR cameras. The drone's movement is governed by a PID velocity controller. The example of the swarm formation control of Crazyflie 2.1 drones is shown in Fig. \ref{fig:real_setup}.

\section{Flocking Methodology}

\subsection{Point Destitution}
The SDF derived from the LLM requires a transformation into a format that is interpretable by the controller. To achieve this, we selected point clouds as our method of choice. However, a simple sampling of points from the SDF's surface presents a challenge, as it may result in the loss of object information at lower sample rates. This is due to the potential proximity of some generated points and the distance of others, leading to an uneven distribution. To address this issue, we implemented a four-step optimization process to ensure a uniform distribution of points on the SDF object's surface. Fig. \ref{fig:point_distribution} illustrates the point distribution visualization.

\begin{figure}[tb]
  \centering
  \includegraphics[width=0.8\columnwidth]{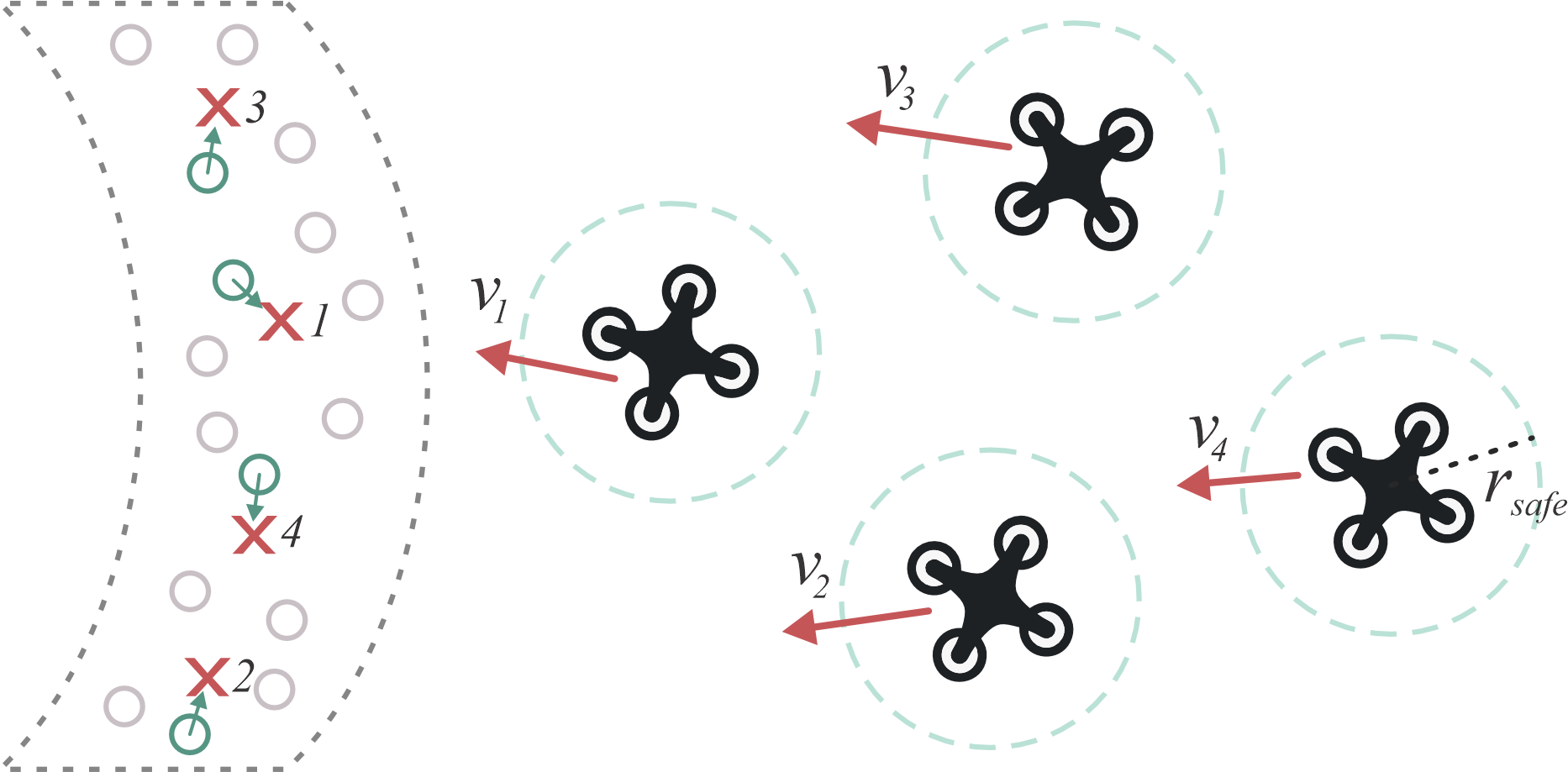}
  \caption{The diagram illustrates the process of drone allocation on the virtual surface. Green dashed circles show the minimum repulsion radius $r_{safe}$, which is utilized to avoid collisions between the drones. Red arrows represent the velocity vectors $v_i$, directing the UAVs towards the surface defined by SDF. Gray circles on this surface are the points randomly generated before optimization. Green circles are the points optimized for our given number of UAVs. Red crosses indicate the positions recalculated for the drones taking into account their flocking behavior.}
  \label{fig:flocking_surface}
  \vspace{-0.3cm}
\end{figure}

At the first stage of our approach, we generate a dense point cloud representation of the object by sampling a substantial number of points within the surface of the SDF. Following this, we employ the L-BFGS-B optimization algorithm to project these points onto the SDF's surface. To accommodate the number of drones in the swarm, it is necessary to downsize the resultant point cloud. To accomplish this, we apply the k-means algorithm, which effectively condenses the point cloud to the required number of points while offering a reliable initial estimation for the optimization algorithm. Finally, we obtain an array of points $P = \left \{ p_1, p_2, ..., p_n \right \}$. We define each point as:
\begin{equation}
\label{eq:point}
p_i = \begin{bmatrix}
x_i\\ 
y_i\\ 
z_i
\end{bmatrix},
\end{equation}
where $x_i$, $y_i$, and $z_i$ are the Cartesian coordinates of each point in the point array. These points serve as the initial guess for the object representation. 

To optimally distribute these points within the SDF, we formulate an optimization problem with the following cost function:
\begin{equation}
\label{eq:cost}
C = \alpha c_{sdf} + \beta c_{dist} + \gamma c_{vol},
\end{equation}
\begin{equation}
\label{eq:sdf_cost}
c_{sdf} = \sum_{i=0}^{n}|f_{sdf}(p_i)| ,
\end{equation}
\begin{equation}
\label{eq:sdf_cost}
c_{dist} = \sum_{i=0}^{n} min(|p_j - p_i|) : j\in [0, n], j\in N, j \neq i ,
\end{equation}
\begin{equation}
\label{eq:sdf_cost}
c_{vol} = V(P) ,
\end{equation}
where $c_{sdf}$ is the cost associated with a point not being located on the SDF's surface; $c_{dist}$ is the distance cost and it is calculated as the sum of the minimum distances between points; $c_{vol}$ is the volume of the convex hull generated by the points $P$. The coefficients $\alpha$, $\beta$, and $\gamma$ are used to balance the contributions of each component to the overall cost function. 

By optimizing the cost function, given in (\ref{eq:cost}) with the L-BFGS-B algorithm we obtain the goal points for the swarm.

\subsection{Flocking Algorithm}

Upon the generation of object points by the point distributor, the control algorithm is initiated to direct the drones to their respective positions as illustrated in Fig.~\ref{fig:flocking_surface}. The drones are sequentially assigned to the closest points in the point cloud. Subsequently, the velocity is computed for each drone based on its current position and target position, leveraging the APF algorithm, as follows:

\begin{equation}
\label{eq:control_velocity}
v = v_{g} + v_{s},
\end{equation}
where $v_{g}$ is the velocity towards the goal position. It is defined by:

\begin{equation}
\label{eq:goal_velocity}
v_{g} = c_{g} (p_{gi} - p_{i}),
\end{equation}
where $p_{gi}$ is the goal position of the $i$-th drone, $p_i$ is the current position of the $i$-th drone, and $c_{g}$ is the scaling coefficient.

The separation velocity $v_{s}$ considers all drones within a sphere of a specified minimum radius and it is given by:

\begin{equation}
\label{eq:separation_velocity}
v_{s} = - c_{s} \sum_{k=0}^{j}(p_k - p_i),
\end{equation}
where $p_k$ is the position of the $k$-th drone within the sphere, $p_i$ is the position of the $i$-th drone, and $c_{s}$ is the scaling coefficient.

\section{Simulation Environment} 
\begin{figure}[tb]
  \centering
  \includegraphics[width=\columnwidth]{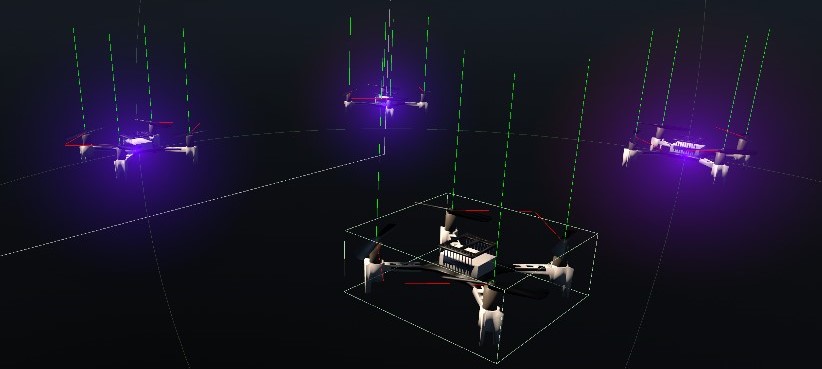}
  \caption{Simulated swarm of drones.}
  \label{fig:simulation_drone}
\end{figure}

A simulation setup was developed in Unity to conduct figure recognition experiments with a large number of drones in a swarm. In this simulation, 3D models of Crazyflie drones with real physical parameters were developed. Each drone is controlled by sending a velocity vector as a control value. To broadcast the speed control vector in the movement of the drone, a PID controller was implemented. The PID controller converts the speed into forces (green line in Fig.~\ref{fig:simulation_drone}) acting on the drone's propellers to implement an accurate simulation of the drone's behavior. 

\begin{figure}[tb]
  \centering
  \includegraphics[width=\columnwidth]{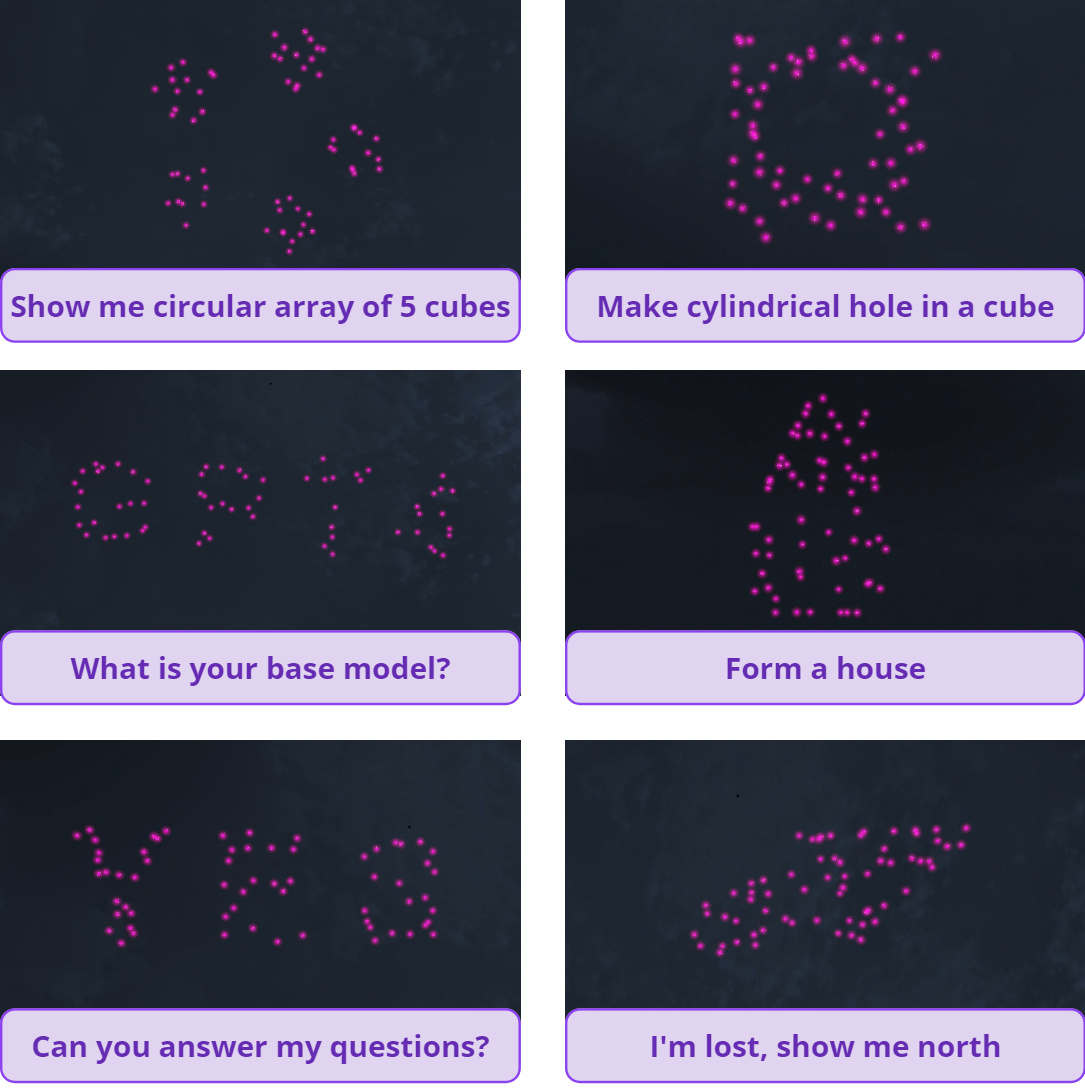}
  \caption{Examples of executing various commands by a swarm of 64 UAVs in the Unity simulation.}
  \label{fig:examples}
\end{figure}

Thus, the Unity environment sends the positions of all drones using the ROS-TCP-Connector to a server with an algorithm for real drones described above, which, after calculating the target speeds, sends them back to the simulation as control values. Fig. \ref{fig:examples} presents examples of the system's operation in forming the geometry of a swarm of 64 drones based on natural language user queries in the Unity simulation. Specifically, it explores the iterative construction of shapes using a combination of primitives, the generation of complex objects based on their names, and engaging in dialogue with users by displaying text responses in the sky.

\subsection{Setup for UAV flock control}

Our setup for controlling real drones includes a swarm of mini-drones Crazyflie 2.1. These quadcopters have advanced features, e.g., self-stabilization and altitude hold, essential for precise and stable flight operations. The ground control station is deployed on a base PC, enabling seamless communication between the modules of the UAV system. This base PC is intricately linked to a Motion capture (mocap) positioning system, enhancing the precision and accuracy of our aerial maneuvers and data collection processes.

\section{User Study of Generated Flock Patterns}


\emph{Participants}: We invited 10 participants (three females, mean age 24.7, SD = 1.22), to experience the visual interpretation of six different geometrical primitives performed by the LLM-driven swarm. Five of the participants had not commonly worked with drones, while two participants had interacted with drones several times. The participants were informed about the experimental procedure and agreed to the consent form. 

\emph{Procedure}: The experimental procedure applied for recognition of swarm-generated figures is based on the methodology suggested by Baza et al. \cite{Baza_2022} for accessing virtual avatar expressive gestures from Russell’s Circumplex model performed with the avatar consisting of a swarm of drones. We have evaluated the recognition rate of five geometric primitives: sphere (SPH), cube (CUBE), tetrahedron (TET), cylinder (CYL), cone (CONE), and one complex pattern of a chess pawn (PAWN). Users watched a series of figures generated through the LLM interface in random order, each figure was demonstrated three times (18 times total).
 
Users were not influenced by other external factors such as the sound or color of the drones. After watching the swarm shaping different patterns, each user was asked to recognize what figure was generated by the flock.

\emph{Experimental Results}:

\begin{table}[!ht]
\centering
\caption{Confusion Matrix of Swarm Shape Recognition}
\label{tab:force}
\begin{tabular}{| c | c | c | c | c | c | c |}
\hline
\multicolumn{1}{|c|}{\%} &\multicolumn{6}{c|}{Estimated}\\
\hline
\textbf{}Performed& SPH & CUBE & TET & CYL & CONE & PAWN\\
\hline
SPH & \textbf{77.8} & \textbf{22.2} &0.0& 0.0 &0.0 &0.0 \\\hline
CUBE & 7.41 & \textbf{92.6} &0.0& 0.0 &0.0 &0.0 \\\hline
TET &0.0 & 3.7 &\textbf{92.6}&0.0& 3.7&0.0 \\\hline
CYL &0.0&0.0&0.0 &\textbf{88.9}&0.0&11.1 \\\hline
CONE & 3.7&0.0&0.0 &\textbf{22.2}& \textbf{63.0}&11.1 \\\hline
PAWN &0.0 & 11.1&11.1&14.8&0.0 &\textbf{63.0} \\\hline
\end{tabular}
\end{table}

To evaluate the statistical significance of the differences between the perception of the flock patterns performed by the simulated swarm, we analyzed the results using a single factor repeated-measures ANOVA, with a chosen significance level of $\alpha < 0.05$. The open-source statistical package Pingouin was used for the statistical analysis. According to the ANOVA results, there is a statistically significant difference in the recognition rates for the different flock patterns, $F(5,54) = 2.58, p = 0.031$. The paired t-tests with one-step Bonferroni correction did not show statistically significant differences between patterns of CYL and CONE ($p_{corr}=0.60$), CYL and PAWN ($p_{corr}=0.36$), CONE and PAWN ($p_{corr}=0.68$), while in all other cases the t-tests showed a statistically significant difference ($p_{corr} < 0.05$). The open-source statistical package Pingouin was used for the statistical analysis. The mean recognition rate of all flock patterns was $80\%$, while a maximum of $93\%$ was achieved for CUBE and TET patterns. The least recognizable patterns were CONE and PAWN ($63\%$ each). We hypothesize that the low recognition rate of these flock patterns was caused by similar shapes of the geometrical primitives and the wide variety of surface parameters generated through LLM.


All participants invited to the flock pattern recognition experiment were then invited to generate different flock patterns in the simulation environment through the FlockGPT interface. After the interaction with FlockGPT, each participant was asked to complete the NASA Task Load Index (NASA-TLX) \cite{NASA} and User Experience \cite{UEQ} questionnaires to assess the pragmatic and hedonic qualities of the interface. 
To pass the unweighted NASA-TLX survey, the participants provided feedback on the following questions: \\
\textbf{Mental Demand:} How much mental and perceptual activity was required (e.g. deciding, calculating, etc.)? Was the task easy or demanding, simple or complex? (Low - High) \\
\textbf{Physical Demand:} How much physical activity was required? Was the task slack or strenuous? (Low - High) \\
\textbf{Temporal Demand:} How much time pressure did you feel due to the pace at which the tasks or task elements occurred? Was the pace slow or rapid? (Low - High) \\
\textbf{Overall Performance:} How successful were you in performing the task? How satisfied were you with your performance? (Perfect - Failure) \\
\textbf{Effort:} How hard did you have to work (mentally and physically) to accomplish your level of performance? (Low - High) \\
\textbf{Frustration Level:} How stressed and annoyed versus relaxed and complacent did you feel during the task? (Low - High)
   
\begin{figure}[t]
\centering
\includegraphics[width=\linewidth]{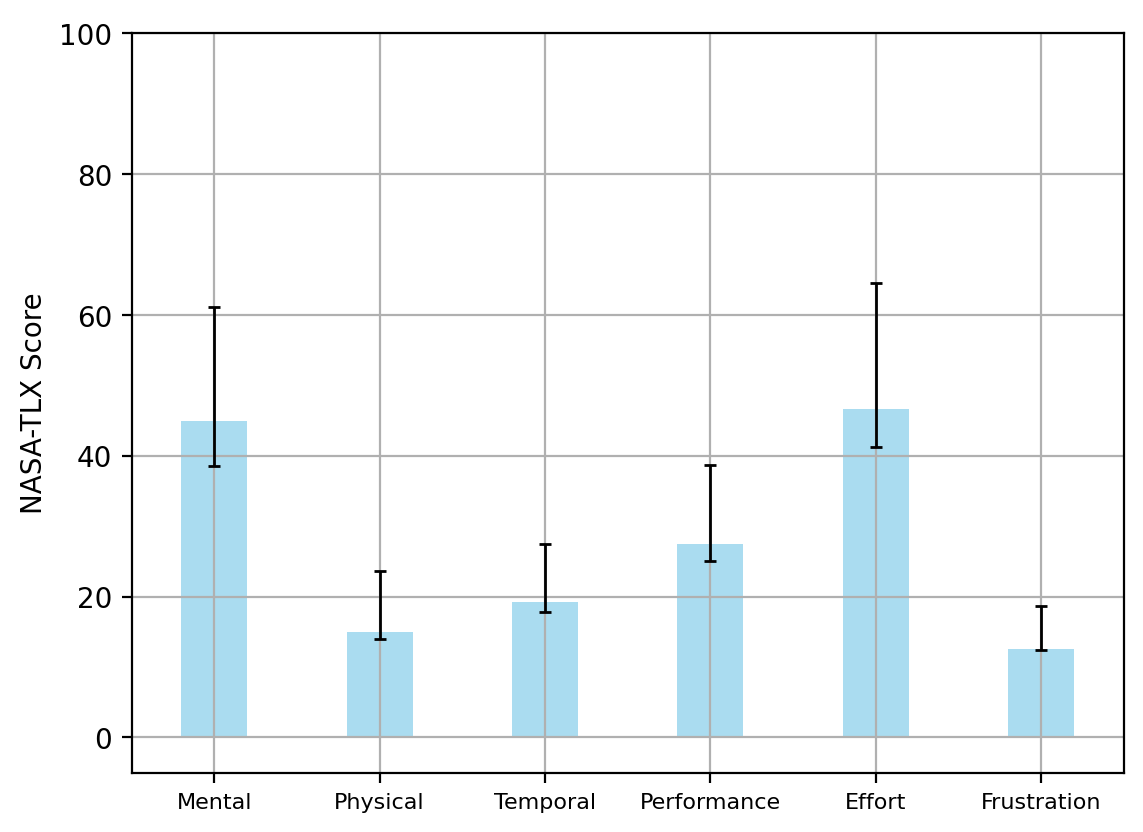}
\caption{Subjective feedback on the 100-point NASA-TLX survey.} \label{fig:likert}
\end{figure}

The results of the unweighted NASA-TLX evaluation score by the study group are shown in Fig.~\ref{fig:likert}. The values for each dimension were calculated on a 20-point Likert scale and mapped to a score from 0 to 100. Users commented on low temporal demand (M = 19.2, SD = 10.6), high performance (M = 26, SD = 13.2, 0 score assessed is perfect), and evaluated mental demand level as medium (M = 45.0, SD = 14.4). The results suggest that future research should focus on developing an input interface with a higher visibility of the pattern before swarm performance. 

\begin{figure}[t]
\centering
\includegraphics[width=\linewidth]{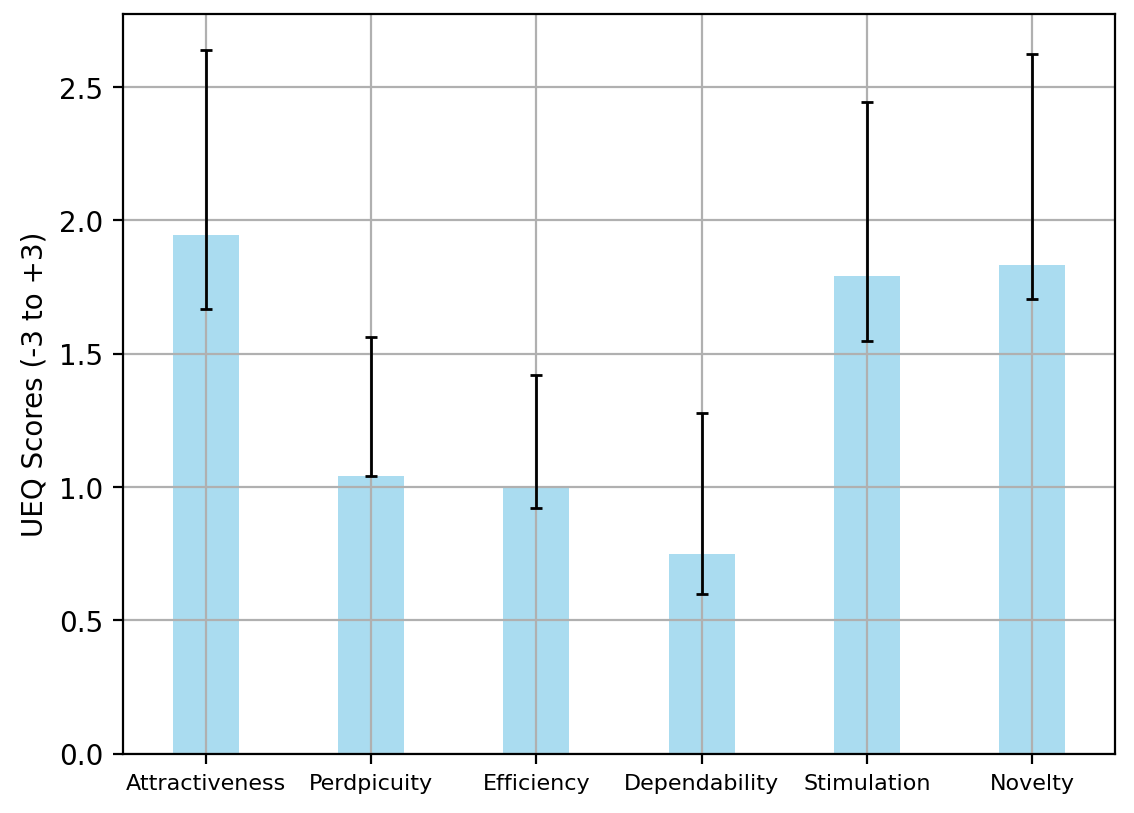}
\caption{Subjective feedback on the 7-point UEQ scale.} \label{fig:likert2}
\end{figure}

The results of the UEQ survey are shown in Fig.~\ref{fig:likert2}. The participants highly rated the stimulation (M = 1.79, SD = 0.71) and novelty (M = 1.83, SD = 0.72) metrics of the system. Moreover, the attractiveness (M = 1.94, SD = 0.74), and hedonic quality (M = 1.81, SD = 0.8) metrics of the developed system were evaluated with high value, suggesting that the developed FlockGPT provided a positive emotional experience to the users. 

\section{Conclusion and Future Work}

The paper presents FlockGPT, the world's first system for managing a highly scalable swarm of drones using intuitive natural language input from the user. LLM made it possible to generate the desired swarm geometry through the signed distance function format coded in Python. Additionally, the system supports user dialogue by generating text-based representations of swarm behavior and allowing for geometry editing in real-time based on clarifying comments. Developed flocking algorithms in swarm control ensure smooth, safe, and optimal transitions between target states. The approach was tested with both a large drone swarm (of 64 drones) in a Unity simulation and a smaller drone swarm (of 8 drones) with real Crazyflie 2.1 mini-drones.

The user study results revealed a high recognition rate for six different flock patterns generated through the LLM-based interface of FlockGPT and simulated by a swarm of drones, with a mean recognition rate of 80\% and a maximum of 93\% for cube and tetrahedron patterns. The least recognizable flock patterns were cone and chess pawn, likely due to their similar shapes and the high variety of the LLM-generated parameters. Users evaluated the developed system using NASA-TLX and UEQ scores, indicating low temporal demand (19.2 score in NASA-TLX), high performance (26 score in NASA-TLX), attractiveness (1.94 UEQ score), and hedonic quality (1.81 UEQ score) of the developed system. Additionally, users noted a medium level of mental demand (45.0 score in NASA-TLX), suggesting that future research should focus on developing a more human-centered input interface. However, most participants highly rated the stimulation (1.79 UEQ score) and novelty (1.83 UEQ score) of the system even with the existing command-line interface.

 FlockGPT can potentially have a strong impact on the intelligence of drone shows, simulation of flocks in VR, and Human-Flock Interaction. Future work will be devoted to the generation of dynamic shapes, e.g. rings of Saturn that can orbit around the planet or completing unfinished buildings with dynamic architectural designs. Moreover, the swarm-based messenger enabling chart through the signs and text in the night sky suggests a new way of communication between people. The path of the drones can be also governed by the emotions of the cue. For example, when the user says “Construct a beautiful house” the FloakGPT will design the shape with extra focus on colors, appearance, and aesthetics.

\bibliographystyle{abbrv-doi-hyperref}

\bibliography{lib}
\addtolength{\textheight }{-6pt}

\appendix 







\end{document}